\documentclass[twoside,11pt,final]{article}

\usepackage{jmlr2e-mine}

\usepackage{color,ifdraft,calc}
\usepackage[T1]{fontenc}
\usepackage{enumitem}
\setlist{nolistsep}

\usepackage{amsmath,amsfonts,amssymb}
\let\qedhere\relax %
\usepackage{xcolor,listings}
\usepackage{booktabs,rotating}

\usepackage[final]{hyperref}  

\newcommand{\liquidSVM}{\texttt{liquidSVM}}
\newcommand{\R}{\texttt{R}}
\newcommand{\Java}{\texttt{Java}}
\newcommand{\Matlab}{\texttt{MATLAB}}
\newcommand{\Python}{\texttt{Python}}
\newcommand{\Spark}{\texttt{Spark}}
\newcommand{\Cxx}{\texttt{C++}}
\newcommand{\Windows}[1][]{\texttt{Windows#1}}
\newcommand{\Mac}{\texttt{MacOS X}}

\newcommand{\libsvm}{\texttt{libsvm}}
\newcommand{\kernlab}{\texttt{kernlab}}
\newcommand{\klaR}{\texttt{klaR}}
\newcommand{\SVMlight}{\texttt{SVMlight}}
\newcommand{\eloll}{\texttt{e1071}}

\newcommand{\BudgetedSVM}{\texttt{BudgetedSVM}}
\newcommand{\EnsembleSVM}{\texttt{EnsembleSVM}}
\newcommand{\GURLS}{\texttt{GURLS}}

\newcommand{\AVX}{\texttt{AVX}}

\newcommand{\AVXtwo}{\texttt{AVX2}}
\newcommand{\SSEtwo}{\texttt{SSE2}}

\newcommand{\Reg}{\textsuperscript\textregistered}
\newcommand{\TM}{\textsuperscript\texttrademark}

\newcommand{\descript}[1]{\paragraph{#1.}}

\newcommand{\g}{\gamma}

\newcommand{\lb}{\lambda}
\newcommand{\x}{\xi}

\newcommand{\snorm}[1]{\Vert #1 \Vert}
\newcommand{\fDlg}{f_{D,\lb,\g}}

\newcommand{\bl}{\left\langle}
\newcommand{\br}{\right\rangle}
\def\<#1>{\bl\;#1\;\br}

\let\phi\varphi
\let\epsilon\varepsilon

\newcommand{\nolabel}[1]{\notag\relax}

\newcommand{\newcl}[1]{\expandafter\def\csname cl#1\endcsname{\mathcal{#1}}}
\newcommand{\newbf}[1]{\expandafter\def\csname bf#1\endcsname{\mathbf{#1}}}
\newcommand{\newds}[1]{\expandafter\def\csname ds#1\endcsname{\mathbb{#1}}}
\newcommand{\newfr}[1]{\expandafter\def\csname fr#1\endcsname{\mathfrak{#1}}}
\newcommand{\newmathop}[1]{\expandafter\DeclareMathOperator\csname #1\endcsname{#1}}
\newcommand{\newaxiom}[1]{\expandafter\def\csname #1\endcsname{\textnormal{\textsc{#1}}}}

\newds{N}
\newds{R}
\newds{Z}
\newds{P}
\newds{C}
\newds{B}
\newds{T}
\newds{D}
\newds{G}
\newds{Q}
\newds{F}

\newcl{A}
\newcl{B}
\newcl{C}
\newcl{D}
\newcl{E}
\newcl{H}
\newcl{L}
\newcl{P}
\newcl{F}
\newcl{G}
\newcl{K}
\newcl{M}
\newcl{I}
\newcl{L}
\newcl{N}
\newcl{M}
\newcl{Q}
\newcl{S}
\newcl{T}

\newbf{R}
\newbf{m}
\newbf{s}
\newbf{A}
\newbf{B}
\newbf{C}
\newbf{w}
\newbf{a}
\newbf{b}
\newbf{M}
\newbf{D}
\newbf{v}
\newbf{c}
\newbf{d}
\newbf{Q}
\newbf{q}
\newbf{y}
\newbf{g}
\newbf{z}
\newbf{i}

\newmathop{Tr}
\newmathop{diag}
\newmathop{Var}
\newmathop{Cov}
\newmathop{diam}
\newmathop{supp}
\newmathop{id}

\newcommand{\qed}{\hfill\rule{7pt}{7pt}}

\newcommand{\refreshQED}{\gdef\qedhere{\gdef\qedhere{\relax}%
\ifmmode\tag*{\qed}\else\qed\fi}\let\myqedhere\qedhere}

\newlength{\myleftmargin}
\newlength{\fixboxwidth}
\jmlrheading{1}{2016}{1-5}{4/00}{10/00}{Ingo Steinwart, and Philipp Thomann}

\ShortHeadings{liquidSVM: A Fast and Versatile SVM package}{Steinwart and Thomann}
\firstpageno{1}

\begin{document}

\title{liquidSVM: A Fast and Versatile SVM package}

\author{\name Ingo Steinwart \email ingo.steinwart@mathematik.uni-stuttgart.de\\
	\name Philipp Thomann \email philipp.thomann@mathematik.uni-stuttgart.de\\
       \addr Institute for Stochastics and Applications \\
       University of Stuttgart, Germany}

\editor{}

\maketitle

\begin{abstract}%
\liquidSVM\ is a   package written in C++ that 
provides SVM-type solvers for various classification and regression tasks.
Because of a fully integrated hyper-parameter selection, very carefully implemented solvers,
multi-threading and  GPU support,
and several built-in data decomposition strategies
 it provides unprecedented speed
for small training sizes as well as for data sets of tens of millions of samples.
Besides the C++ API and a command line interface, bindings to  \R, \Matlab, \Java, \Python, and \Spark\ are available.
We present a brief description of the package and report 
experimental comparisons to other SVM packages.
\end{abstract}

\begin{keywords}
C++, Support Vector Machine, non-parametric classification, non-parametric regression, CUDA, open source, R, Java, MATLAB, Python, Spark
\end{keywords}


\section{Introduction}

Support vector machines (SVMs) and related kernel-based learning algorithms 
are a well-known class of machine learning algorithms, for which a couple of 
very popular implementations such as \SVMlight\
\cite{Joachims99a}
and \libsvm\
\cite{ChLi11a}
as well as some recent   packages for large-scale data sets \cite{tacchetti13a,djuric13a,claesen14a}
 already exist. 
Despite this, training SVMs is still relatively costly,
in particular if it comes to very large data sets and/or hyper-parameter selection.
In addition, most packages do not include solvers for more involved estimation
problems such as quantile/expectile regression or classification with a constraint on the false alarm 
rate. The goal of  \liquidSVM, which is licensed under 
AGPL 3.0, is to address these issues.
In a nutshell, the key features of \liquidSVM\ are:
\begin{itemize}[noitemsep]
	\item Fully integrated hyper-parameter selection based on cross validation
	\item Extreme speed on both small and large data sets
	\item Inclusion of a variety of different classification and regression scenarios 
	\item Good default values and high flexibility for experts
	\item Flexible user interface ranging from a \Cxx\ API and a command line version
to bindings for \R, \Matlab, \Java, \Python, and \Spark
\end{itemize}
The main software package can be obtained from
\url{http://www.isa.uni-stuttgart.de/software}
and the bindings are packaged under the respective directories
\href{http://www.isa.uni-stuttgart.de/software/R/}{\texttt{/R}},
\href{http://www.isa.uni-stuttgart.de/software/matlab/}{\texttt{/matlab}},
\href{http://www.isa.uni-stuttgart.de/software/java/}{\texttt{/java}},
\href{http://www.isa.uni-stuttgart.de/software/python/}{\texttt{/python}}, and
\href{http://www.isa.uni-stuttgart.de/software/spark/}{\texttt{/spark}},
resp.
\liquidSVM\ has been tested on several versions of 
\verb!Linux! and \Mac, as well as on \Windows[ 8].
For the latter two systems pre-compiled binaries are provided, too.


\section{Software Description}

In  \liquidSVM\ an application cycle is divided into a training phase,
in which various SVM models are created and validated, a selection phase, in which
the SVM models that best satisfy a certain criterion are selected, and a test phase,
in which the selected models are applied to test data. These three phases are based upon 
several components, which can be freely combined. In the following, we briefly describe 
the four most important components:

\descript{Solvers} The solvers create  SVM models $\fDlg$ by solving 
\begin{equation}\label{svm-problem}
   \fDlg = \arg\min_{f\in H_\g} \lb \snorm f_{H_\g}^2 + \frac 1 n \sum_{i=1}^n L_w\bigl(y_i, f(x_i)\bigr)\, .
\end{equation}
Here $\lb>0$ is a regularization parameter, $H_\g$ is a reproducing kernel Hilbert space with kernel $k_\g$
and kernel parameter 
$\g>0$,
$D=((x_1,y_1),\dots,(x_n,y_n))$ is a labeled data set, and $L_w$ is a loss function with weight parameter $w>0$.
Currently,  \liquidSVM\ include solvers for the (weighted) hinge loss used for classification,
the least squares loss used for mean regression, the pinball loss used for quantile regression, and the 
asymmetric least squares loss used for expectile regression. Moreover, the standard Gaussian RBF kernel 
and the Laplacian kernel are implemented and it is possible to add own normalized kernels.

\descript{Hyper-Parameter Selection}
The  problem \eqref{svm-problem} has the free parameters 
$\lb$ and $\g$, and in some situations, also $w$.
\liquidSVM\ automatically determines good values for these parameters by performing 
$k$-fold cross validation (CV) over an (adaptive) grid of candidate values.
The user can choose between different fold generation methods 
and can also determine the candidate grid and the
loss function used on the validation fold.
In addition, the user can decide, whether one SVM model or $k$ SVM models are created 
during the selection phase and, if applicable, how these $k$ models are combined during the test phase.
To speed up the CV, the required kernel matrices may be re-used
and all solvers contain advanced warm start options. By default, the most time efficient combinations are 
picked.

\descript{Managing Working Sets}
Some learning scenarios such as one-versus-all (OvA) multiclass classification
require to solve \eqref{svm-problem} for a couple of different subsets $D$ of the full 
data set. In \liquidSVM\ each such data set is associated to a \emph{task}.
Moreover, a well-known strategy to speed up training is to split 
the data into smaller parts or \emph{cells}, see e.g.~\cite{BoVa92a,VaBo93a}.
Currently,  \liquidSVM\ offers to create tasks according to
OvA, AvA, as well as to weighted classification and quantile/expectile regression.
In addition, several methods to create random or spatially defined cells are implemented.
Different task and cell creation methods can be freely combined   and at the end, hyper-parameter 
selection as described above is performed on each resulting cell.

\descript{User Interfaces and Pre-defined Learning Scenarios}
Besides the \Cxx\ class API 
 for experienced users,
\liquidSVM\ also has a command line interface (CLI) as well as bindings to 
\R, \Matlab, \Python, and \Java.
Both the CLI and the bindings
contain routines for various standard learning scenarios such as:
(weighted) binary classification,  multiclass classification (both AvA and OvA),
Neyman-Pearson-type classification, least squares regression, quantile regression, and 
expectile regression. These routines also have a simplified interface to facilitate 
a fast and easy access to  the functionality of \liquidSVM.


%
%
%
%
%
\section{Implementation Details}

\liquidSVM\ is written in \Cxx\ 
and its main functionality is accessible 
through a small number of high-level \Cxx\ classes. 
The code is divided into four parts: 
\emph{a)} SVM independent code for I/O-operations, data set manipulations, and generic $k$-fold CV, 
\emph{b)} SVM related code such as the core solvers,
\emph{c)} code for some extra CLI tools,
and \emph{d)} code related to the bindings.

The routines for 
computing the kernel matrices and for evaluating the SVM models 
on the test data 
are parallelized
to run on multiple cores and for \verb!Linux!, \verb!Cuda! implementations
of these routines 
do also exist.
Time critical inner loops may be vectorized 
with the following instruction sets:
\verb!SSE2!, \verb!AVX!, and \verb!AVX2!.
When compiling \liquidSVM\ under \verb!Linux! or \Mac\ the best instruction set for the current machine is chosen, 
while in the bindings we additionally offer an explicit compilation with \verb!SSE2!.


All currently available solvers are based on the design principles 
for the hinge loss solver
described by \cite{StHuSc11a}. For the least squares and quantile solver,
the corresponding modifications were straightforward, while for the expectile 
solver more care was necessary, see \cite{FaSt17a}.

All bindings share a common C-interface to \liquidSVM's \Cxx\ code.
They perform all operations in-memory and in a single process, which may control several threads.



\section{Benchmarks}

To illustrate the speed of \liquidSVM\ we report some comparisons to other 
available implementations.
Here, we only give a brief summary of our results, more extensive experiments 
as well as further details can be found in the appendix
in Section~\ref{sec:appendix-benchmarks}.
Except in the comparison to \GURLS, we only 
considered binary classification since this is
the common denominator of the considered implementations.
In our experiments we performed 5-fold CV to select the hyper-parameters from a
$10\!\times\!11$-grid suggested by \libsvm,
and for \liquidSVM\ we  additionally considered its default  $10\!\times\!10$-grid.
For packages not containing   a CV routine, we manually implemented it by  wrapping  loops.

For small data sets of size $n=4000$ we 
considered three implementations that have an \R-interface, namely:
package \eloll, which binds to \libsvm,
package \klaR~\citep{weihs05}, which wraps \SVMlight, and
package \kernlab~\citep{karatzoglou04}, which is implemented entirely in \R.
The corresponding results, which are summarized in Table~\ref{table:traintimeCV},
show that even with a single thread \liquidSVM\  is more than an order of
magnitude faster. 
Similarly, Table~\ref{table:GURLS} shows that \liquidSVM\ is between  7 and 35 times faster than
\GURLS~\citep{tacchetti13a} on four multiclass data sets of size $n\leq 10000$.

For medium-sized data sets with $n\leq 280000$ we considered the following two  implementations
that allow a partition of the training set into cells:
\BudgetedSVM~\citep{djuric13a}   and \EnsembleSVM~\citep{claesen14a}. 
Both have a  parameter $k$, which can be compared to our cell size.
In Table~\ref{table:combinedLarge} we report the results for 
 $k=1000$ (results for  $k=500, 3000$ can be found in the appendix).
It turns out that \liquidSVM\ is in many cases two orders of magnitude  faster, and in most cases it also achieves  a significant reduction 
of the test error.

Finally, for large training sets up to around 30 million samples we performed experiments on a \Spark\ cluster, see Table~\ref{table:spark}. 
Here, we actually achieved in most cases a slightly super-linear speed-up
compared to the single-node run,
since 
less overhead was created.



\newcommand{\theVspace}{\vspace{-.25cm}}

\begin{table}[htp]
\centering
\begingroup\footnotesize
\scalebox{0.8}{
\begin{tabular}{lrrrrrrrr}
  \toprule
 & \texttt{dim} & \texttt{liquidSVM} & \texttt{(libsvm grid)} & \texttt{(sec.)} & \texttt{(outer cv)} & \texttt{libsvm} & \texttt{kernlab} & \texttt{SVMlight} \\ 
  \midrule
\textsc{bank-marketing} & 16 & $\times 0.4 $ & $\times 1 $ & 27.4s & $\times 11.2 $ & $\times 19.1 $ & $\times 32.3 $ & $\times 235.3 $ \\ 
  \textsc{cod-rna} & 8 & $\times 0.6 $ & $\times 1 $ & 17.1s & $\times 14.7 $ & $\times 24.9 $ & $\times 27.3 $ & $\times 460.1 $ \\ 
  \textsc{covtype} & 55 & $\times 0.4 $ & $\times 1 $ & 35.4s & $\times 11.5 $ & $\times 34.1 $ & $\times 52.3 $ & $\times 615.6 $ \\ 
  \textsc{thyroid-ann} & 21 & $\times 0.4 $ & $\times 1 $ & 32.2s & $\times 9.8 $ & $\times 12.7 $ & $\times 26.5 $ & $\times 244.1 $ \\ 
   \bottomrule
\end{tabular}
}
\endgroup

\theVspace
\caption{\small Cross validation time for small (n=4000) data sets.
The times are given relative to our fully optimized  implementation 
on the hyper-parameter grid of \libsvm,
for which we also present  the absolute time in seconds.
Times are averaged over 10 independent repetitions.
\liquidSVM\ \texttt{(outer cv)} uses \eloll\texttt{::tune}
and solves in every grid-point a single SVM.
\texttt{SVMlight} is quite slow here due to  disk accesses in the wrapper. In these experiments
\liquidSVM\  is single-threaded, the other implementations do not support multi-threading.
The test errors are comparable (see Table~\ref{table:errCV-all} in the Appendix).%
} 
\label{table:traintimeCV}
\end{table}

\begin{table}[htp]
\centering
\begingroup\footnotesize
\scalebox{0.8}{
\begin{tabular}{lrrr|rrr|rrr}
  \toprule
 &&&number of& \multicolumn{3}{c|}{time (sec.)} & \multicolumn{2}{c}{classification error (in \%)} &\\
 &size&dim&classes& \texttt{Our} & \texttt{GURLS} & factor & \texttt{Our} & \texttt{GURLS} &  \texttt{GURLS} paper \\
  \midrule
\textsc{optdigit}   & 3823 & 64 & 10 &  47.1 &  348.5 & $\times$ 7.4 &  1.17 &  1.11 & 1.7 \\ 
\textsc{landsat}    & 4435 & 36 &  6 &  53.5 &  540.0 & $\times$10.1 &  7.75 & 14.45 & 9.6 \\ 
\textsc{pendigit}   & 7494 & 16 & 10 & 167.7 & 2281.6 & $\times$13.6 &  1.46 &  1.63 & 1.6 \\ 
\textsc{covtype}   & 10000 & 54 &  2 & 135.7 & 4748.0 & $\times$35.0 & 16.00 & 18.30 & --- \\ 
   \bottomrule
\end{tabular}
}
\endgroup
\theVspace
\caption{\small Comparison to \GURLS\ for multi-class classification.
For our implementations we used OvA with the least-squares solver and no cell splitting.
\GURLS\ has an internal parameter selection for the cost parameter
while the kernel parameter was set by their heuristic involving the lower quartile of the distance matrix values.
Both implementations used full multi-threading
(6 physical for \liquidSVM, 12 logical for \GURLS).%
}
\label{table:GURLS}
\end{table}

\begin{table}[htp]
\centering
\begingroup\footnotesize
\scalebox{0.7}{
\begin{tabular}{lrr|rrrrrr|rrrrr}
  \toprule
 & \texttt{size} & \texttt{dim} & \texttt{liquidSVM} & \texttt{(sec.)} & \texttt{(libsvm grid)} & \texttt{Overlap} & \texttt{Bsvm} & \texttt{Esvm} & \texttt{liquidSVM} & \texttt{(libsvm grid)} & \texttt{Overlap} & \texttt{Bsvm} & \texttt{Esvm} \\ 
  \midrule
\textsc{covtype} & 10000 & 54 & $\times 1.0 $ & 11.7s & $\times 0.7 $ & $\times 2.4 $ & $\times 521.8 $ & $\times 40.2 $ & 16.67 & 17.03 & 16.60 & 19.90 & 21.86 \\ 
  \textsc{covtype} & 40000 & 54 & $\times 1.0 $ & 50.2s & $\times 0.7 $ & $\times 5.9 $ & $\times 445.8 $ & $\times 49.3 $ & 10.80 & 13.23 & 10.29 & 19.34 & 19.80 \\ 
  \textsc{covtype} & 100000 & 54 & $\times 1.0 $ & 117.6s & $\times 0.7 $ & $\times 8.4 $ & $\times 430.0 $ & $\times 87.8 $ & 7.97 & 10.91 & 7.09 & 18.41 & 19.12 \\ 
  \textsc{ijcnn1} & 49990 & 23 & $\times 1.0 $ & 39.1s & $\times 1.0 $ & $\times 8.1 $ & $\times 549.2 $ & $\times 49.6 $ & 1.91 & 1.65 & 1.16 & 1.62 & 6.45 \\ 
  \textsc{webspam} & 280000 & 255 & $\times 1.0 $ & 374.5s & $\times 1.0 $ & $\times 91.8 $ & $\times 407.8 $ & $\times 474.5 $ & 1.63 & 1.80 & 0.93 & 2.61 & 4.11 \\ 
   \bottomrule
\end{tabular}
}
\endgroup

\theVspace
\caption{\small Benchmarks for splitting mid-sized data sets with cell size 1000.
The left side presents single-threaded training times (relative to \liquidSVM) including 5-fold CV,
and on the right-hand side the corresponding classification errors (in \%) can be found.
\texttt{Overlap} uses our solver  but with overlapping  instead of mutually disjoint cells.
Our errors are almost always significantly better, while in many cases the speed-up is two 
orders of magnitudes.
}
\label{table:combinedLarge}
\end{table}

%

\begin{table}[htp]
\centering
\begingroup\footnotesize
\scalebox{0.8}{
\begin{tabular}{lrr|rrr|rr}
  \toprule
 &&& \multicolumn{2}{c}{time (min.)} & & \multicolumn{2}{c}{classification error (in \%)} \\
 & size & dim & distributed & single node & speedup & distributed & single node \\
  \midrule
\textsc{covtype} & 464429 & 54 & 1.5 & 8.8 & 5.9 & 4.17 & 4.34 \\ 
  \textsc{susy} & 4499999 & 18 & 8.8 & 133.6 & 15.2 & 26.38 & 22.95 \\ 
  \textsc{hepmass} & 7000000 & 28 & 12.4 & 267.9 & 21.6 & 15.09 & 14.82 \\ 
  \textsc{higgs} & 9899999 & 28 & 23.2 & 368.4 & 15.9 & 32.87 & 32.67 \\ 
  \textsc{ecbdl} & 29418893 & 631 & 449.1 & --- & --- & 1.85 & --- \\ 
%
   \bottomrule
\end{tabular}
}
\endgroup
\theVspace
\caption{\small Benchmarks on a Spark cluster with 14 workers, each using 6 threads.
The data was first split into coarse cells of size approximately 20000 and
every cell was collected on a single worker.
Then each such coarse cell was solved locally using fine cells of size at most 2000.
The single node experiments are taken from \citet{large-scale-svm} using the command line version on the same machines using 6 threads.
The speedup is in most cases super-linear since less overhead is created.
All times include 5-fold CV on our $10\!\times\!10$-grid.%
}
\label{table:spark}
\end{table}


\bibliography{steinwart-books,steinwart-proc,steinwart-article,steinwart-mine,others}

\clearpage
\appendix

\clearpage

\section{Usage}
\label{sec:appendix-bindings}

\definecolor{bluekeywords}{rgb}{0.13,0.13,1}
\definecolor{greencomments}{rgb}{0,0.5,0}
\definecolor{redstrings}{rgb}{0.9,0,0}

\lstset{
showspaces=false,
showtabs=false,
breaklines=true,
showstringspaces=false,
breakatwhitespace=true,
escapeinside={(*@}{@*)},
commentstyle=\color{greencomments},
keywordstyle=\color{bluekeywords}\bfseries,
stringstyle=\color{redstrings},
basicstyle=\footnotesize\ttfamily
}

\subsection{Command Line Interface}
\lstset{language=bash}

For Linux download \href{http://www.isa.uni-stuttgart.de/software/liquidSVM.tar.gz}{\texttt{liquidSVM.tar.gz}} and use the following commands to compile and
train and test  a multi-class SVM on our banana data set
with display verbosity 1 and using 2 threads:
\begin{lstlisting}
tar xzf liquidSVM.tar.gz
cd liquidSVM
make all
cd scripts
./mc-svm.sh banana-mc 1 2
\end{lstlisting}
%

\subsection{\R}
\lstset{language=R}

You can use the following example:
\begin{lstlisting}
install.packages("liquidSVM", repos="http://www.isa.uni-stuttgart.de/R")
library(liquidSVM)
d <- liquidData('banana-mc')  # load the multi-class banana data set
model <- mcSVM( Y ~ . , d$train, display=1, threads=2)
result <- test(model, d$test)
\end{lstlisting}
%
More information can be found in the \href{http://www.isa.uni-stuttgart.de/software/R/demo.html}{demo} and \href{http://www.isa.uni-stuttgart.de/software/R/documentation.html}{documentation} vignettes online
Also consider the manuals for the package \verb!?liquidSVM! or for the commands
\verb!?lsSVM!, \verb!?mcSVM!, etc.

\subsection{\Java}
\lstset{language=Java}

For installation download \href{http://www.isa.uni-stuttgart.de/software/java/liquidSVM-java.zip}{\texttt{liquidSVM-java.zip}} and unzip it.
The classes are all in package \verb!de.uni_stuttgart.isa.liquidsvm! and an easy example is:
\begin{lstlisting}
SVM s = new LS(trainX, trainY, new Config().display(1).threads(2));
ResultAndErrors result = s.test(testX, testY);
\end{lstlisting}

If this is implemented in file \verb!Example.java! this can be compiled and run using
\begin{lstlisting}[language=bash]
javac -classpath liquidSVM.jar Example.java
java -Djava.library.path=. -cp .:liquidSVM.jar Example
\end{lstlisting}

\subsection{\Matlab\ / \texttt{Octave}}
\lstset{language=Matlab}

For installation download the Toolbox
\href{http://www.isa.uni-stuttgart.de/software/matlab/liquidSVM.mltbx}%
{\texttt{liquidSVM.mltbx}} and install it in \Matlab\ by double clicking it.
Then you can use it like:
\begin{lstlisting}
makeliquidSVM native
load data
model = mcSVM(banana_mc_train_x, banana_mc_train_y, 'DISPLAY','1','THREADS','2');
model.test(banana_mc_test_x, banana_mc_test_y);
\end{lstlisting}
If the training labels are \verb!categorical! they are
transparently converted to integer labels
and in this case, if no learning scenario is specified,
binary or multi-class classification is performed by default.
The code also works in \texttt{Octave} if you use
\href{http://www.isa.uni-stuttgart.de/software/matlab/liquidSVM-octave.zip}%
{\texttt{liquidSVM-octave.zip}}.

\subsection{\Python}
\lstset{language=Python}

Install using
\begin{lstlisting}[language=bash]
pip install --user \
  http://www.isa.uni-stuttgart.de/software/python/liquidSVM-python.tar.gz
\end{lstlisting}
this also will install \verb!numpy! if it is not available yet.
Then in \Python\ the package is used as:
\begin{lstlisting}
from liquidSVM import *
model = mcSVM(iris, iris_labs, display=1,threads=2)
result, err = model.test(iris, iris_labs)
\end{lstlisting}

\subsection{\Spark}
\lstset{language=Scala}

Download \href{http://www.isa.uni-stuttgart.de/software/spark/liquidSVM-spark.zip}{\texttt{liquidSVM-spark.zip}} and unzip it 
and issue:
\begin{lstlisting}[language=bash]
make lib
export LD_LIBRARY_PATH=.:$LD_LIBRARY_PATH
$SPARK_HOME/bin/spark-submit --master local[*] \
  --class de.uni_stuttgart.isa.liquidsvm.spark.liquidSVMsparkApp \
  liquidSVM-spark.jar covtype.10000
\end{lstlisting}
If you have configured \Spark\ to be used on a cluster
with \texttt{Hadoop} use:
\begin{lstlisting}[language=bash]
hdfs dfs -put data/covtype-full.train.csv data/covtype-full.test.csv .
make lib
$SPARK_HOME/spark-submit --files ../libliquidsvm.so \
  --conf spark.executor.extraLibraryPath=. \
  --conf spark.driver.extraLibraryPath=. \
  --class de.uni_stuttgart.isa.liquidsvm.spark.liquidSVMsparkApp \
  --num-executors 14 liquidSVM-spark.jar covtype-full
\end{lstlisting}

\section{Details and further benchmarks}
\label{sec:appendix-benchmarks}

In this section we present more extensive benchmarks and give some technical details.
First consider Table~\ref{table:implementations-overview} for an overview of
all considered implementations.

Let us first give some general details on the hyper-parameter grids we used
for cross-validation.
By \libsvm\ grid we mean here the $10\!\times\!11$ grid given by
\begin{align*}
  \gamma&\in\{\,2^3,2,2^{-1},2^{-3},2^{-5},2^{-7},2^{-9},2^{-11},2^{-13},2^{-15}\,\},\\
  \text{cost}&\in\{\,2^{-5},2^{-3},2^{-1},2,2^{3},2^{5},2^{7},2^{9},2^{11},2^{13},2^{15}\,\},
\end{align*}
which is suggested by the default values in the file \texttt{tools/grid.py} of \libsvm.
For \liquidSVM\ we performed experiments both on this grid and
on our default $10\!\times\!10$ geometrically spaced hyper-parameter grid
where the endpoints are scaled to accommodate
the number of samples in every fold, the cell size, and the dimension.

\begin{table}[htp]
\centering
\begingroup\footnotesize
\scalebox{0.52}{
\let\x\checkmark
\begin{tabular}{lp{2.6cm}cccccccp{4cm}cp{1.75cm}p{4.5cm}}
  \toprule
 &  Loss & Parallel & GitHub & CLI & MATLAB & R  & C++ & cv & kernels & RBF-definition & Data Format & MLOSS paper data sets \\
 \midrule
\libsvm & (weighted) Hinge, $\epsilon,\nu$ & (CUDA) & \x & \x & \x & \x & \x & \x & linear, polynomial, RBF, sigmoid & $\exp(-\gamma\|u-v\|^2)$ & libsvm & \\
\kernlab & (weighted) Hinge, $\epsilon,\nu$, etc. &  &  &  &  & \x &  & & linear, polynomial, RBF, sigmoid, user-defined, etc. & $\exp(-\gamma\|u-v\|^2)$  & \R &\\
\SVMlight & (weighted) Hinge, $\epsilon$, ranking & & & \x & \x & \x & \x & & linear, polynomial, RBF, sigmoid, user-defined & $\exp(-\gamma\|u-v\|^2)$ & libsvm & \\
\midrule
\EnsembleSVM & Hinge  & \x & \x & \x &  &  & &(\x) & linear, polynomial, RBF, sigmoid, user-defined, etc. & $\exp(-\gamma\|u-v\|^2)$ & libsvm, csv, sparse csv & \sc covtype, ijcnn1 \\
\BudgetedSVM & Hinge &  & sf.net & \x & \x & & & & linear, polynomial, RBF, sigmoid, exponential, user-defined & $\exp(-\gamma\|u-v\|^2)$ & libsvm & \sc webspam, rcv1, mnist8m-bin \\
\GURLS & LS & \x & \x &  & \x &  & \x & $\lambda$ & && & \sc optdigit, landsat, pendigit, letter, isolet \\
\midrule
\liquidSVM & (weighted) Hinge, (asymmetric) LS, Pinball & \x + CUDA & \em soon & \x  & \x & \x & \x &\x
& RBF, Poisson, user-defined & $\exp(-\frac{\|u-v\|^2}{\g^2})$  & libsvm, csv, uci  & \sc covtype, ijcnn1, webspam, optdigit, landsat, pendigit, bank-marketing, cod-rna,thyroid-ann, susy, hepmass, higgs, ebcdl \\
   \bottomrule
\end{tabular}
}
\endgroup
\caption{\small Features of the different implementations we consider here.
In the top are the implementations we used for small-sized data sets,
in the middle those for mediums-sized, and in the bottom our.%
}
\label{table:implementations-overview}
\end{table}

\subsection{Small-sized data sets}
Every data set was split into 10 training sets of size $n\in\{1000, 2000, 4000\}$ and 10 test sets.
Based on the training a scaling was determined and both training and test set were normalized by that.
We did all the computation in R and used
package \eloll\ which binds to \libsvm,
package \klaR\ which wraps \SVMlight, and
package \kernlab\ which is implemented in \R.
Unfortunately, \klaR\ wraps around the command line version of \SVMlight\ and hence
has to write all data to disk, which gives a big performance penalty.
We used an Intel\Reg\ Core\TM\ i7-4770 cpu at 3.40GHz with 16GB memory and \AVXtwo\ running Debian Linux.
For the results consider Tables~\ref{table:traintimeCV-all} and~\ref{table:errCV-all}

\begin{table}[htp]
\centering
\footnotesize
$n=1000$:\ 
\begingroup\footnotesize
\scalebox{0.8}{
\begin{tabular}{lrrrrrrr}
  \toprule
 & \texttt{dim} & \texttt{liquidSVM} & \texttt{(libsvm grid)} & \texttt{(outer cv)} & \texttt{libsvm} & \texttt{kernlab} & \texttt{SVMlight} \\ 
  \midrule
\textsc{bank-marketing} & 16 & 0.8 & 1.7 & 52.9 & 43.4 & 53.9 & 3057.8 \\ 
  \textsc{cod-rna} & 8 & 0.7 & 1.2 & 47.1 & 34.5 & 36.7 & 665.6 \\ 
  \textsc{covtype} & 55 & 1.0 & 1.9 & 73.5 & 87.7 & 98.8 & 8607.0 \\ 
  \textsc{thyroid-ann} & 21 & 0.9 & 1.9 & 57.5 & 34.9 & 48.0 & 3027.2 \\ 
   \bottomrule
\end{tabular}
}
\endgroup
\\[1ex]
$n=2000$:\ 
\begingroup\footnotesize
\scalebox{0.8}{
\begin{tabular}{lrrrrrrr}
  \toprule
 & \texttt{dim} & \texttt{liquidSVM} & \texttt{(libsvm grid)} & \texttt{(outer cv)} & \texttt{libsvm} & \texttt{kernlab} & \texttt{SVMlight} \\ 
  \midrule
\textsc{bank-marketing} & 16 & 2.9 & 6.3 & 119.2 & 142.3 & 215.1 & 4649.3 \\ 
  \textsc{cod-rna} & 8 & 2.6 & 4.4 & 105.2 & 118.5 & 126.6 & 1609.1 \\ 
  \textsc{covtype} & 55 & 3.4 & 7.8 & 156.4 & 319.8 & 412.2 & 14495.8 \\ 
  \textsc{thyroid-ann} & 21 & 3.0 & 7.3 & 119.9 & 107.0 & 174.9 & 5005.0 \\ 
   \bottomrule
\end{tabular}
}
\endgroup
\\[1ex]
$n=4000$:\ 
\begingroup\footnotesize
\scalebox{0.8}{
\begin{tabular}{lrrrrrrr}
  \toprule
 & \texttt{dim} & \texttt{liquidSVM} & \texttt{(libsvm grid)} & \texttt{(outer cv)} & \texttt{libsvm} & \texttt{kernlab} & \texttt{SVMlight} \\ 
  \midrule
\textsc{bank-marketing} & 16 & 11.1 & 27.4 & 306.2 & 522.9 & 884.5 & 6450.0 \\ 
  \textsc{cod-rna} & 8 & 10.1 & 17.1 & 251.8 & 425.1 & 466.3 & 7870.2 \\ 
  \textsc{covtype} & 55 & 13.5 & 35.4 & 407.6 & 1206.8 & 1853.1 & 21803.2 \\ 
  \textsc{thyroid-ann} & 21 & 11.5 & 32.2 & 315.1 & 409.1 & 853.4 & 7849.8 \\ 
   \bottomrule
\end{tabular}
}
\endgroup

\caption{\small Cross validation time for different implementations on small-sized data sets.
Times are means of 10 independent repetitions and all involve 5-fold cross-validation on a 10x11 grid -- only our optimized version uses its usual 10x10 grid.
\liquidSVM\ \texttt{(outer cv)} uses \eloll\texttt{::tune}
and solves in every grid-point a single SVM.
\texttt{SVMlight} is quite slow here due to need of disk access in the wrapper.
\liquidSVM\ here is single-threaded, the others do not support multi-threading
} 
\label{table:traintimeCV-all}
\end{table}

\begin{table}[htp]
\centering
\footnotesize
$n=1000$:\ 
\begingroup\footnotesize
\scalebox{0.8}{
\begin{tabular}{lrrrrr}
  \toprule
 & \texttt{liquidSVM} & \texttt{(libsvm grid)} & \texttt{libsvm} & \texttt{kernlab} & \texttt{SVMlight} \\ 
  \midrule
\textsc{bank-marketing} & 0.1115 & 0.1118 & 0.1110 & 0.1113 & 0.1153 \\ 
  \textsc{cod-rna} & 0.0446 & 0.0454 & 0.0434 & 0.0452 & 0.0497 \\ 
  \textsc{covtype} & 0.2526 & 0.2474 & 0.2580 & 0.2540 & 0.2506 \\ 
  \textsc{thyroid-ann} & 0.0534 & 0.0511 & 0.0488 & 0.0477 & 0.0460 \\ 
   \bottomrule
\end{tabular}
}
\endgroup
\\[1ex]
$n=2000$:\ 
\begingroup\footnotesize
\scalebox{0.8}{
\begin{tabular}{lrrrrr}
  \toprule
 & \texttt{liquidSVM} & \texttt{(libsvm grid)} & \texttt{libsvm} & \texttt{kernlab} & \texttt{SVMlight} \\ 
  \midrule
\textsc{bank-marketing} & 0.1090 & 0.1085 & 0.1086 & 0.1087 & 0.1160 \\ 
  \textsc{cod-rna} & 0.0416 & 0.0431 & 0.0414 & 0.0416 & 0.0485 \\ 
  \textsc{covtype} & 0.2284 & 0.2263 & 0.2315 & 0.2324 & 0.2440 \\ 
  \textsc{thyroid-ann} & 0.0512 & 0.0484 & 0.0457 & 0.0464 & 0.0464 \\ 
   \bottomrule
\end{tabular}
}
\endgroup
\\[1ex]
$n=4000$:\ 
\begingroup\footnotesize
\scalebox{0.8}{
\begin{tabular}{lrrrrr}
  \toprule
 & \texttt{liquidSVM} & \texttt{(libsvm grid)} & \texttt{libsvm} & \texttt{kernlab} & \texttt{SVMlight} \\ 
  \midrule
\textsc{bank-marketing} & 0.1070 & 0.1066 & 0.1065 & 0.1065 & 0.1165 \\ 
  \textsc{cod-rna} & 0.0401 & 0.0413 & 0.0391 & 0.0393 & 0.0482 \\ 
  \textsc{covtype} & 0.2038 & 0.2055 & 0.2074 & 0.2067 & 0.2411 \\ 
  \textsc{thyroid-ann} & 0.0457 & 0.0458 & 0.0429 & 0.0433 & 0.0803 \\ 
   \bottomrule
\end{tabular}
}
\endgroup

\caption{\small Mean classification errors for different implementations on small-sized data sets.%
} 
\label{table:errCV-all}
\end{table}

We also reproduced some of the results of \GURLS~\citep{tacchetti13a}.
As they do not provide a command line interfaces, we had to adapt their example C++ program
to use their library in Version~2.0.
After communication with the authors we used their internal parameter selection for the cost parameters
and their heuristic for the kernel parameter (lower quartile of distances).
Our reproduced runtimes using their software are about factor 40 faster than their reported ones and
we attribute this to the following factors:
it seems their written times are reported for MATLAB where we used their C++ version,
we used 6 threads, and our cpu was clocked 37\% higher.

\subsection{Medium-sized data sets}

We considered several implementations that use a random chunk approach to scale SVMs.
Based on the training data a scaling was determined and both training and test set were normalized by that.
Then we used Bash scripts to perform cross validation 
We used an Intel\Reg\ Core\TM\ i7-3930K cpu at 3.20GHz with 64GB memory and \AVX\ running Debian Linux.
The results for cell sizes $k=500,3000$
are in Tables~\ref{table:trainLarge-all} and~\ref{table:errLarge-all}
(and we also repeat here those for $k=1000$).

It is commendable, that \BudgetedSVM\ publish on their
web page\footnote{\url{http://www.dabi.temple.edu/budgetedsvm/docs/run\_budgetedsvm\_algs.m}} the concrete hyper-parameters
they were using for experiments  (at least for $k=500$).
However they not describe, how they found them.
Our CV found for \textsc{webspam} the same cost $0.125$,
however we found $\gamma=8$ to have better validation error than their $\gamma=16$
(which is in their definition of the kernel function actually $\tilde\gamma=32$).
We also tried out their BSGD-variant, however (at least for $k=500$)
this was always 10 times slower than their LLSVM-variant and hence would need much more time
to finish.
We used Version~1.1.

\EnsembleSVM\ (version 2.0) has a nice interface to mask sets for training and validation splits.
Sadly, there is no out of the box script to perform cross-validation only
an example on their homepage.

A bit annoyingly, on the $10\!\times\!11\!\times\!5$ grid training was hanging some times,
and one had to kill the process
(after holidays we found out that on solver had run without convergence for several days).
To remain comparable, we subtracted the whole excess time!

\begin{table}[htp]
\centering
\footnotesize
\begin{tabular}{ll}
$k=500$:\\
\begingroup\footnotesize
\scalebox{0.8}{
\begin{tabular}{lrrrrr|rrrr}
  \toprule
 & \texttt{liquidSVM} & \texttt{(libsvm grid)} & \texttt{Overlap} & \texttt{Bsvm} & \texttt{Esvm} & \texttt{liquidSVM} & \texttt{(libsvm grid)} & \texttt{Overlap} & \texttt{Esvm} \\ 
  \midrule
\textsc{covtype.10000} & 0.1 & 0.1 & 0.5 & 33.7 & 17.4 & 0.1 & 0.1 & 0.5 & 9.0 \\ 
  \textsc{covtype.40000} & 0.6 & 0.4 & 4.1 & 138.4 & 37.8 & 0.5 & 0.4 & 4.3 & 28.9 \\ 
  \textsc{covtype.100000} & 1.4 & 0.9 & 11.8 & 305.1 & 94.0 & 1.2 & 0.9 & 11.4 & 65.1 \\ 
  \textsc{ijcnn1} & 0.4 & 0.4 & 9.9 & 140.0 & 27.7 & 0.4 & 0.4 & 11.7 & 24.1 \\ 
  \textsc{webspam} & 4.7 & 4.9 & 293.5 & 997.5 & 775.4 & 4.3 & 4.5 & 275.1 & 754.0 \\ 
   \bottomrule
\end{tabular}
}
\endgroup
\\[3ex]
$k=1000$:\\
\begingroup\footnotesize
\scalebox{0.8}{
\begin{tabular}{lrrrrr|rrrr}
  \toprule
 & \texttt{liquidSVM} & \texttt{(libsvm grid)} & \texttt{Overlap} & \texttt{Bsvm} & \texttt{Esvm} & \texttt{liquidSVM} & \texttt{(libsvm grid)} & \texttt{Overlap} & \texttt{Esvm} \\ 
  \midrule
\textsc{covtype.10000} & 0.2 & 0.1 & 0.5 & 102.1 & 7.9 & 0.2 & 0.1 & 0.3 & 4.3 \\ 
  \textsc{covtype.40000} & 0.8 & 0.6 & 4.9 & 372.9 & 41.2 & 0.6 & 0.4 & 3.1 & 26.3 \\ 
  \textsc{covtype.100000} & 2.0 & 1.4 & 16.5 & 842.8 & 172.1 & 1.4 & 0.9 & 9.9 & 136.6 \\ 
  \textsc{ijcnn1} & 0.7 & 0.6 & 5.3 & 358.3 & 32.4 & 0.4 & 0.5 & 2.9 & 23.9 \\ 
  \textsc{webspam} & 6.2 & 6.2 & 572.9 & 2545.4 & 2961.5 & 3.9 & 4.0 & 295.0 & 2663.4 \\ 
   \bottomrule
\end{tabular}
}
\endgroup
\\[3ex]
$k=3000$:\\
\begingroup\footnotesize
\scalebox{0.8}{
\begin{tabular}{lrrrrr|rrrr}
  \toprule
 & \texttt{liquidSVM} & \texttt{(libsvm grid)} & \texttt{Overlap} & \texttt{Bsvm} & \texttt{Esvm} & \texttt{liquidSVM} & \texttt{(libsvm grid)} & \texttt{Overlap} & \texttt{Esvm} \\ 
  \midrule
\textsc{covtype.10000} & 0.4 & 0.3 & 0.8 & 690.4 & 343.4 & 0.2 & 0.1 & 0.3 & 83.6 \\ 
  \textsc{covtype.40000} & 1.7 & 1.3 & 5.1 & 1711.0 & 388.1 & 0.8 & 0.4 & 2.0 & 122.0 \\ 
  \textsc{covtype.100000} & 4.2 & 3.5 & 29.7 & 3513.3 & 518.6 & 1.8 & 1.1 & 11.0 & 250.7 \\ 
  \textsc{ijcnn1} & 1.6 & 1.7 & 2.7 & 1841.2 & 190.1 & 0.5 & 0.6 & 0.8 & 81.5 \\ 
  \textsc{webspam} & 12.6 & 12.4 & 2519.7 & 12393.8 & 3052.7 & 4.4 & 4.3 & 683.7 & 2370.5 \\ 
   \bottomrule
\end{tabular}
}
\endgroup
\\[3ex]
\end{tabular}
\caption{\small Cross validation time (in min.) for medium-sized data sets.
In the left part the times are single-threaded, in the right part, they are with 6 threads.
For $k=3000$ the others only use a $9\!\times\!10$-grid%
.%
} 
\label{table:trainLarge-all}
\end{table}

\begin{table}[htp]
\centering
\footnotesize
\begin{tabular}{ll}
$k=500$:&
\begingroup\footnotesize
\scalebox{0.8}{
\begin{tabular}{lrrrrr}
  \toprule
 & \texttt{liquidSVM} & \texttt{(libsvm grid)} & \texttt{Overlap} & \texttt{Bsvm} & \texttt{Esvm} \\ 
  \midrule
\textsc{covtype.10000} & 16.78 & 17.04 & 15.84 & 21.96 & 22.43 \\ 
  \textsc{covtype.40000} & 11.52 & 13.78 & 10.19 & 20.89 & 21.93 \\ 
  \textsc{covtype.100000} & 8.27 & 11.12 & 7.22 & 21.08 & 21.65 \\ 
  \textsc{ijcnn1} & 2.82 & 2.10 & 1.21 & 2.17 & 6.96 \\ 
  \textsc{webspam} & 1.77 & 1.93 & 0.97 & 3.04 & 5.48 \\ 
   \bottomrule
\end{tabular}
}
\endgroup
\\[1ex]
$k=1000$:&
\begingroup\footnotesize
\scalebox{0.8}{
\begin{tabular}{lrrrrr}
  \toprule
 & \texttt{liquidSVM} & \texttt{(libsvm grid)} & \texttt{Overlap} & \texttt{Bsvm} & \texttt{Esvm} \\ 
  \midrule
\textsc{covtype.10000} & 16.67 & 17.03 & 16.60 & 19.90 & 21.86 \\ 
  \textsc{covtype.40000} & 10.80 & 13.23 & 10.29 & 19.34 & 19.80 \\ 
  \textsc{covtype.100000} & 7.97 & 10.91 & 7.09 & 18.41 & 19.12 \\ 
  \textsc{ijcnn1} & 1.91 & 1.65 & 1.16 & 1.62 & 6.45 \\ 
  \textsc{webspam} & 1.63 & 1.80 & 0.93 & 2.61 & 4.11 \\ 
   \bottomrule
\end{tabular}
}
\endgroup
\\[1ex]
$k=3000$:& 
\begingroup\footnotesize
\scalebox{0.8}{
\begin{tabular}{lrrrrr}
  \toprule
 & \texttt{liquidSVM} & \texttt{(libsvm grid)} & \texttt{Overlap} & \texttt{Bsvm} & \texttt{Esvm} \\ 
  \midrule
\textsc{covtype.10000} & 16.58 & 16.46 & 16.31 & 17.10 & 17.31 \\ 
  \textsc{covtype.40000} & 10.76 & 12.82 & 10.60 & 15.66 & 16.22 \\ 
  \textsc{covtype.100000} & 7.72 & 10.74 & 7.30 & 14.72 & 15.71 \\ 
  \textsc{ijcnn1} & 1.34 & 1.11 & 1.38 & 1.12 & 3.64 \\ 
  \textsc{webspam} & 1.35 & 1.48 & 0.84 & 2.23 & 3.06 \\ 
   \bottomrule
\end{tabular}
}
\endgroup
\\[1ex]
\end{tabular}
\caption{\small Classification errors (in \%) for different implementations on medium-sized data sets.
\BudgetedSVM\ (LLSVM) published for \textsc{webspam} error rates of $3.46, 2.60,$ and $1.99$, resp.\ \citep{djuric13a}.
For \EnsembleSVM\  published for \textsc{covtype} and \textsc{ijcnn1} error rates of $11\%$, and $9\%$ resp.\ \citep{claesen14a}.
For $k=3000$ the others only use a $9\times10$-grid%
.%
} 
\label{table:errLarge-all}
\end{table}

\subsection{Large-sized data sets}

As architecture, we used Intel\Reg\ Xeon\Reg\ CPUs (E5-2640 0 at 2.50GHz, May 2013) with Ubuntu Linux.
There were two NUMA-sockets each with a CPU having 6 physical cores with 128GB memory and \AVX.
We thank the Institute of Mathematics at the University of Zurich
for providing us access to those machines.

Let us repeat here our description from \cite{large-scale-svm}:
The data set was saved on a Hadoop distributed file system on one master and 7 worker machines of the above type.
In a first step, the data was split into coarse cells of estimated size 20000 by the following procedure.
A subset of the training data was sampled and sent to the master machine
where 300--8000 centres were found and these centres were sent back to the worker machines.
Now each worker machine could assign locally to every of it's samples the coarse cell
in the Voronoi sense.
Finally a \Spark-shuffle was performed: Every cell was assigned to one of the workers
and all its samples were sent to that worker.

In the second step every such coarse cell--now being on one physical machine--was
used for training by our C++ implementation discussed above:
this in particular means that each coarse cell was again split into fine cells of size 2000.
Obviously this now was done in parallel on all worker nodes.
The test set was also split into the coarse cells
and then by our implementation further into fine cells for prediction.

The single node times and errors (as well as the above two-paragraph description) are from \citet{large-scale-svm}.
For these the command line interface was used on the same machines.
Hence they have much more overhead in terms of disk access and retraining.
This explains the super-linear speed-up, even though for the \Spark-version
there has to be done some shuffling over an Ethernet-LAN.

\section{Benchmarks for liquidSVM configurations}

\newcommand{\BenchmarkOur}[1]{
\def\arch{#1}
\input{benchmarks/tables/timeOur-1000-\arch.tex}
\input{benchmarks/tables/timeOur-2000-\arch.tex}
\input{benchmarks/tables/timeOur-4000-\arch.tex}
\input{benchmarks/tables/timeOur-6000-\arch.tex}
}

\liquidSVM\ can be configured extensively.
They are described in the documentation of the software.
We selected some of those and performed experiments for the small-sized data sets.
The selection is for the following parameters:
\begin{description}
\item[\texttt{threads}:] Controls the number of threads in the kernel evaluations
\item[\texttt{grid\_choice=0}:] our standard $10\!\times\!10$ hyper-parameter grid.
\item[\texttt{grid\_choice=1}:] a $15\!\times\!15$ hyper-parameter grid
\item[\texttt{grid\_choice=2}:] a $20\!\times\!20$ hyper-parameter grid
\item[\texttt{adaptivity\_control:}] selects adaptively a subset of the hyper-parameter grid
\item[\texttt{voronoi:}] value $5$ uses a overlapping decomposition into cells,
value $6$ specifies our recursive partitioning scheme.
The optional second parameter specifies the maximal cell size (default 2000).
\end{description}
The training times and classification errors are given in Tables~\ref{table:combinedOur-1000-native}--\ref{table:combinedOur-6000-native}
There are more configurations that are useful to control time and memory consumption
however we did not benchmark them here.

\begin{table}[htp]
\centering
\begingroup\footnotesize
\scalebox{0.8}{
\begin{tabular}{lrrrr|rrrr}
  \toprule
 & \begin{sideways} \textsc{bank-marketing} \end{sideways} & \begin{sideways} \textsc{cod-rna} \end{sideways} & \begin{sideways} \textsc{covtype} \end{sideways} & \begin{sideways} \textsc{thyroid-ann} \end{sideways} & \begin{sideways} \textsc{bank-marketing} \end{sideways} & \begin{sideways} \textsc{cod-rna} \end{sideways} & \begin{sideways} \textsc{covtype} \end{sideways} & \begin{sideways} \textsc{thyroid-ann} \end{sideways} \\ 
  \midrule
\texttt{threads=1} & 1.18 & 1.23 & 1.19 & 1.20 & 11.06 & 4.32 & 24.51 & 5.17 \\ 
  \texttt{threads=2} & 1.15 & 1.20 & 1.17 & 1.17 & 11.14 & 4.46 & 24.40 & 5.37 \\ 
  \texttt{threads=3} & 1.06 & 1.05 & 1.05 & 1.06 & 11.04 & 4.44 & 24.87 & 5.32 \\ 
  \texttt{threads=4} & 1.00 & 1.00 & 1.00 & 1.00 & 11.19 & 4.40 & 24.42 & 5.45 \\ 
  \texttt{grid\_choice=1} & 2.36 & 2.15 & 2.45 & 2.52 & 11.13 & 4.36 & 24.41 & 4.93 \\ 
  \texttt{grid\_choice=2} & 7.03 & 5.61 & 7.51 & 8.31 & 11.19 & 4.44 & 24.45 & 4.71 \\ 
  \texttt{adaptivity\_control=1} & 0.89 & 0.88 & 0.86 & 0.84 & 11.09 & 4.44 & 24.40 & 5.32 \\ 
  \texttt{adaptivity\_control=2} & 0.79 & 0.77 & 0.80 & 0.74 & 11.16 & 4.34 & 24.75 & 5.26 \\ 
  \texttt{adaptivity\_control=2, grid\_choice=2} & 3.55 & 2.66 & 2.96 & 3.62 & 11.13 & 4.40 & 24.67 & 5.13 \\ 
  \texttt{voronoi=5} & 0.99 & 0.99 & 1.00 & 1.01 & 11.12 & 4.49 & 24.72 & 5.17 \\ 
  \texttt{voronoi=6} & 0.99 & 1.00 & 0.99 & 1.01 & 11.10 & 4.36 & 24.74 & 5.25 \\ 
  \texttt{voronoi=c(5,1000)} & 0.99 & 0.99 & 0.99 & 1.00 & 11.02 & 4.34 & 24.48 & 5.31 \\ 
  \texttt{voronoi=c(6,1000)} & 1.00 & 1.00 & 1.01 & 1.00 & 11.21 & 4.35 & 25.04 & 5.33 \\ 
   \bottomrule
\end{tabular}
}
\endgroup
\caption{On the left: training time for different configurations relative to \texttt{threads=4}.
                                                     On the right: classification errors in \%. They are all averaged over 10 repetitions (n=1000).} 
\label{table:combinedOur-1000-native}
\end{table}

\begin{table}[htp]
\centering
\begingroup\footnotesize
\scalebox{0.8}{
\begin{tabular}{lrrrr|rrrr}
  \toprule
 & \begin{sideways} \textsc{bank-marketing} \end{sideways} & \begin{sideways} \textsc{cod-rna} \end{sideways} & \begin{sideways} \textsc{covtype} \end{sideways} & \begin{sideways} \textsc{thyroid-ann} \end{sideways} & \begin{sideways} \textsc{bank-marketing} \end{sideways} & \begin{sideways} \textsc{cod-rna} \end{sideways} & \begin{sideways} \textsc{covtype} \end{sideways} & \begin{sideways} \textsc{thyroid-ann} \end{sideways} \\ 
  \midrule
\texttt{threads=1} & 1.67 & 1.81 & 1.59 & 1.64 & 10.84 & 4.11 & 22.83 & 4.94 \\ 
  \texttt{threads=2} & 1.25 & 1.29 & 1.21 & 1.25 & 10.81 & 4.23 & 22.80 & 4.78 \\ 
  \texttt{threads=3} & 1.08 & 1.11 & 1.07 & 1.06 & 10.86 & 4.13 & 22.59 & 4.91 \\ 
  \texttt{threads=4} & 1.00 & 1.00 & 1.00 & 1.00 & 10.84 & 4.31 & 22.78 & 4.73 \\ 
  \texttt{grid\_choice=1} & 2.66 & 2.16 & 2.81 & 2.71 & 10.83 & 4.16 & 22.61 & 4.62 \\ 
  \texttt{grid\_choice=2} & 9.55 & 6.77 & 10.36 & 11.69 & 10.90 & 4.19 & 22.89 & 4.67 \\ 
  \texttt{adaptivity\_control=1} & 0.90 & 0.86 & 0.90 & 0.76 & 10.86 & 4.16 & 22.82 & 4.94 \\ 
  \texttt{adaptivity\_control=2} & 0.76 & 0.76 & 0.78 & 0.66 & 10.90 & 4.13 & 22.59 & 4.97 \\ 
  \texttt{adaptivity\_control=2, grid\_choice=2} & 4.28 & 2.84 & 3.78 & 3.85 & 10.81 & 4.16 & 22.69 & 4.60 \\ 
  \texttt{voronoi=5} & 1.00 & 1.01 & 0.99 & 0.97 & 10.85 & 4.21 & 22.62 & 4.80 \\ 
  \texttt{voronoi=6} & 1.02 & 1.01 & 1.01 & 0.95 & 10.93 & 4.24 & 22.70 & 4.81 \\ 
  \texttt{voronoi=c(5,1000)} & 0.89 & 1.20 & 0.85 & 0.93 & 11.06 & 4.25 & 22.98 & 5.23 \\ 
  \texttt{voronoi=c(6,1000)} & 0.70 & 0.75 & 0.67 & 0.65 & 11.11 & 4.36 & 22.87 & 5.41 \\ 
   \bottomrule
\end{tabular}
}
\endgroup
\caption{On the left: training time for different configurations relative to \texttt{threads=4}.
                                                     On the right: classification errors in \%. They are all averaged over 10 repetitions (n=2000).} 
\label{table:combinedOur-2000-native}
\end{table}

\begin{table}[htp]
\centering
\begingroup\footnotesize
\scalebox{0.8}{
\begin{tabular}{lrrrr|rrrr}
  \toprule
 & \begin{sideways} \textsc{bank-marketing} \end{sideways} & \begin{sideways} \textsc{cod-rna} \end{sideways} & \begin{sideways} \textsc{covtype} \end{sideways} & \begin{sideways} \textsc{thyroid-ann} \end{sideways} & \begin{sideways} \textsc{bank-marketing} \end{sideways} & \begin{sideways} \textsc{cod-rna} \end{sideways} & \begin{sideways} \textsc{covtype} \end{sideways} & \begin{sideways} \textsc{thyroid-ann} \end{sideways} \\ 
  \midrule
\texttt{threads=1} & 1.76 & 1.99 & 1.67 & 1.81 & 10.71 & 4.04 & 20.31 & 4.57 \\ 
  \texttt{threads=2} & 1.20 & 1.25 & 1.19 & 1.24 & 10.67 & 4.12 & 20.33 & 4.63 \\ 
  \texttt{threads=3} & 1.05 & 1.08 & 1.05 & 1.07 & 10.73 & 4.05 & 20.35 & 4.70 \\ 
  \texttt{threads=4} & 1.00 & 1.00 & 1.00 & 1.00 & 10.75 & 4.04 & 20.22 & 4.68 \\ 
  \texttt{grid\_choice=1} & 2.74 & 2.13 & 3.02 & 3.01 & 10.76 & 4.15 & 20.53 & 4.55 \\ 
  \texttt{grid\_choice=2} & 11.11 & 7.23 & 12.38 & 15.03 & 10.65 & 4.05 & 20.52 & 4.57 \\ 
  \texttt{adaptivity\_control=1} & 0.88 & 0.90 & 0.89 & 0.72 & 10.70 & 4.08 & 20.40 & 4.69 \\ 
  \texttt{adaptivity\_control=2} & 0.73 & 0.75 & 0.79 & 0.60 & 10.73 & 4.01 & 20.27 & 4.57 \\ 
  \texttt{adaptivity\_control=2, grid\_choice=2} & 4.08 & 2.67 & 3.45 & 2.92 & 10.66 & 4.16 & 20.61 & 4.52 \\ 
  \texttt{voronoi=5} & 0.71 & 0.90 & 0.74 & 0.76 & 10.83 & 4.06 & 20.57 & 4.65 \\ 
  \texttt{voronoi=6} & 0.45 & 0.50 & 0.49 & 0.44 & 10.84 & 4.06 & 20.47 & 4.87 \\ 
  \texttt{voronoi=c(5,1000)} & 0.66 & 0.74 & 0.53 & 0.85 & 10.86 & 4.03 & 20.57 & 4.77 \\ 
  \texttt{voronoi=c(6,1000)} & 0.38 & 0.44 & 0.35 & 0.35 & 10.92 & 4.09 & 20.74 & 5.06 \\ 
   \bottomrule
\end{tabular}
}
\endgroup
\caption{On the left: training time for different configurations relative to \texttt{threads=4}.
                                                     On the right: classification errors in \%. They are all averaged over 10 repetitions (n=4000).} 
\label{table:combinedOur-4000-native}
\end{table}

\begin{table}[htp]
\centering
\begingroup\footnotesize
\scalebox{0.8}{
\begin{tabular}{lrrrr|rrrr}
  \toprule
 & \begin{sideways} \textsc{bank-marketing} \end{sideways} & \begin{sideways} \textsc{cod-rna} \end{sideways} & \begin{sideways} \textsc{covtype} \end{sideways} & \begin{sideways} \textsc{thyroid-ann} \end{sideways} & \begin{sideways} \textsc{bank-marketing} \end{sideways} & \begin{sideways} \textsc{cod-rna} \end{sideways} & \begin{sideways} \textsc{covtype} \end{sideways} & \begin{sideways} \textsc{thyroid-ann} \end{sideways} \\ 
  \midrule
\texttt{threads=1} & 1.79 & 2.01 & 1.70 &  & 10.55 & 4.00 & 18.86 &  \\ 
  \texttt{threads=2} & 1.19 & 1.25 & 1.17 &  & 10.61 & 3.97 & 18.85 &  \\ 
  \texttt{threads=3} & 1.06 & 1.08 & 1.05 &  & 10.58 & 3.98 & 18.75 &  \\ 
  \texttt{threads=4} & 1.00 & 1.00 & 1.00 &  & 10.54 & 3.98 & 18.57 &  \\ 
  \texttt{grid\_choice=1} & 2.85 & 2.18 & 3.17 &  & 10.56 & 4.03 & 18.75 &  \\ 
  \texttt{grid\_choice=2} & 12.20 & 7.45 & 13.77 &  & 10.66 & 4.00 & 18.63 &  \\ 
  \texttt{adaptivity\_control=1} & 0.88 & 0.91 & 0.87 &  & 10.57 & 4.00 & 18.80 &  \\ 
  \texttt{adaptivity\_control=2} & 0.75 & 0.78 & 0.79 &  & 10.55 & 3.99 & 18.77 &  \\ 
  \texttt{adaptivity\_control=2, grid\_choice=2} & 4.25 & 2.58 & 2.80 &  & 10.59 & 4.01 & 18.73 &  \\ 
  \texttt{voronoi=5} & 0.55 & 0.60 & 0.52 &  & 10.69 & 3.95 & 18.84 &  \\ 
  \texttt{voronoi=6} & 0.32 & 0.34 & 0.31 &  & 10.80 & 3.98 & 18.91 &  \\ 
  \texttt{voronoi=c(5,1000)} & 0.49 & 0.65 & 0.50 &  & 10.79 & 3.94 & 18.94 &  \\ 
  \texttt{voronoi=c(6,1000)} & 0.26 & 0.28 & 0.25 &  & 10.96 & 4.09 & 19.29 &  \\ 
   \bottomrule
\end{tabular}
}
\endgroup
\caption{On the left: training time for different configurations relative to \texttt{threads=4}.
                                                     On the right: classification errors in \%. They are all averaged over 10 repetitions (n=6000).} 
\label{table:combinedOur-6000-native}
\end{table}



\descript{Architecture}
Native code can be compiled using several levels of single-instruction-multiple-data (SMD) instruction sets:
\SSEtwo, \AVX, \AVXtwo.
Compiling with \texttt{native} will select the highest available setting.
All of the experiments up to here were compiled with \texttt{native}.
We invested some effort to also compile the other implementations using \texttt{native}
yet they did not benefit.
For ours we report results for different compiler settings in Tables~\ref{table:timeOur-1000-archs}--\ref{table:timeOur-6000-archs}. Obviously \texttt{native} here gives \AVXtwo.

\def\arch{archs}
\input{benchmarks/tables/timeOur-1000-\arch.tex}
\input{benchmarks/tables/timeOur-2000-\arch.tex}
\input{benchmarks/tables/timeOur-4000-\arch.tex}
\input{benchmarks/tables/timeOur-6000-\arch.tex}


\end{document}